\documentclass[conference]{IEEEtran}
\IEEEoverridecommandlockouts
\usepackage{cite}
\usepackage{amsmath,amssymb,amsfonts}
\usepackage{graphicx}
\usepackage{textcomp}
\usepackage{xcolor}

\usepackage{hyperref}

\usepackage{amsthm}

\newtheorem{theorem}{Theorem}

\usepackage{algorithm}
\usepackage{algpseudocode}

\def\BibTeX{{\rm B\kern-.05em{\sc i\kern-.025em b}\kern-.08em
    T\kern-.1667em\lower.7ex\hbox{E}\kern-.125emX}}
\begin{document}

\title{Towards Evolutionary Optimization Using the Ising Model}

\author{\IEEEauthorblockN{Simon Klüttermann}

simon.kluettermann@cs.tu-dortmund.de}

\maketitle
\begin{abstract}
In this paper, we study the problem of finding the global minima of a given function. Specifically, we consider complicated functions with numerous local minima, as is often the case for real-world data mining losses.

We do so by applying a model from theoretical physics to create an Ising model-based evolutionary optimization algorithm.

Our algorithm creates stable regions of local optima and a high potential for improvement between these regions. This enables the accurate identification of global minima, surpassing comparable methods, and has promising applications to ensembles.
\end{abstract}


\section{Introduction}\label{sec:intro}

The development of gradient descent \cite{gradientdescent} made it possible to create and improve many data mining algorithms, for example, by allowing methods based on neural networks to be trained efficiently. Still, gradient descent methods have their limits. Most importantly, gradient descent requires a continuous loss function and thus cannot be applied to every type of model. While we have historically preferred to describe complicated relations using intricate mathematical formulas, with gradient descent, we must hide these precise mathematical objects within huge neural network matrices. This is not always possible \cite{libschitzNN} and made it almost impossible to understand the underlying relation \cite{surveyexplainability}.

A potential solution to improve this would be to employ more specialized models and optimize them using a different paradigm, such as evolutionary optimization \cite{evolution,evoOpti}. While evolutionary optimization is often much more time-consuming, variants are applicable to almost every minimization task. Neither the loss function nor its inputs need to be continuous, and it can even allow for optimizing more complicated structures, such as lists of variable length or a tree encoding a mathematical function.

However, due to the increased time cost, evolutionary optimization is typically specialized only to find a good enough optimum. While evolutionary approaches are less likely to get stuck in a local optimum than gradient descent methods, we will see later that they are usually not able to even come close to the global optimum.
While this is a good approach, for example, in hyperparameter optimization \cite{flaml}, where function evaluations are very costly, and the performance increase might often be tiny, the situation is arguably different in data mining.
When searching for an explainable model, only locally optimal solutions also represent a flawed explanation of the underlying process, as there is an explanation that describes it more effectively. This also implies that a locally optimal solution generalizes less well to a new dataset.

To test algorithms in this problem situation, we propose a test function in this paper, which features many local minima of varying qualities. These are also distributed in a way that makes it easy to evaluate the number of local minima found.

Using this test situation, we propose a new evolutionary optimization paradigm that is particularly effective in finding global minima.
This algorithm is inspired by the Ising model \cite{isingModell} from statistical physics, which is often studied because it exhibits the emergence of regions with slightly different behavior. We observe that when extending this algorithm to evolutionary optimization, the emergence of regions remains. This enables our optimization algorithm to consider a much wider range of possible solutions, thereby increasing the likelihood of finding the global minimum.

Additionally, we notice that our optimization can be very helpful for ensemble models \cite{ensemble-survey}. While finding the global optimum of a loss function usually implies a valuable data mining model, combining different models into an ensemble can be even more effective. This is because while a single model might make mistakes, these mistakes are often canceled by averaging multiple ones \cite{ensembleerrorcancel}.

However, to do so, we must find multiple significantly different solutions to a given data mining task, as repeating the same errors does not allow them to cancel each other out. This is not necessarily an easy task, as it often relies on different initializations finding different local minima. And even if it works, it requires running a potentially very slow evolutionary optimization algorithm many times.

The diversity of solutions our algorithm provides can solve both problems. Instead of requiring multiple optimizations, we can benefit from different solutions in different regions.

Concretely, our contributions in this paper are threefold:
\begin{itemize}
    \item Definition of an easy-to-study function for the problem of evolutionary global minima optimization.
    \item Suggestion of a new physics-based optimization algorithm.
    \item Evaluation of its properties and benefits.
\end{itemize}

To help with further research, our code is available at \url{https://github.com/psorus/IsingEvo}.

\section{Related Work}\label{sec:rw}
\subsection{Evolutionary Optimization}
The problem of optimization \cite{optimization} is central to machine learning. As a result, there are several ways to address this issue. A common method is Gradient Descent \cite{gradientdescent}, which efficiently trains complicated neural networks. However, Gradient Descent requires a continuous loss function and struggles to handle randomness effectively. Both are conditions that restrict the type of problems that can be optimized. Additionally, Gradient Descent can often not find the global minimum of a loss function, but only a local minimum.

Evolutionary optimization solves some of these problems. Instead of requiring a continuous loss function differential, it alters the current input randomly and either accepts or rejects the changed value depending on whether it decreases the loss function. This allows applying it to both functions of non-numeric values, as well as to loss functions that are not differentiable \cite{evolution}.

Additionally, there are numerous variants of evolutionary optimization algorithms. Notable are Genetic Algorithms
\cite{geneticalgo}, Particle Swarm Optimization \cite{particleswarm}, and Differential Evolution \cite{diffevolution}. A few specific algorithms are described in Sec. \ref{sec:comparison}.

When searching for the global optimum, the optimization algorithm needs to be able to make a locally suboptimal choice. While this is easy to implement for an evolutionary algorithm, by adding a probability to create a suboptimal choice, it is not optimal. Theoretically, it might be possible to go from a local optimum to a different one, but this is still very unlikely, as it often requires choosing the suboptimal choice many times in a row.
Instead, we suggest a method to increase the variance of inputs tried, which makes it considerably more likely to find the global minimum.

\subsection{Ising Model}
We achieve this by utilizing an Ising model-inspired algorithm. This model, studied in 1924 by Ernest Ising in his dissertation \cite{isingModell}, was initially used to model spin interactions: Two neighboring spins are either aligned ($s_i\cdot s_j=1$) or opposite to each other ($s_i\cdot s_j=-1$), and the energy of the system (similar to a loss function) depends on this interaction $E=-\sum_{\text{i neighbour of j}} s_i\cdot s_j$.

It is one of the most commonly studied models in statistical physics, because it is one of the simplest models to show a phase transition \cite{isingreview}. This means its behavior is drastically different based on one control parameter (the temperature $T$ or the more commonly used inverse temperature $\beta=\frac{1}{T}$): For low temperatures, spins are aligned and stable regions form. But for high temperatures, the spins become chaotic.

Notably, this effect also depends on the dimension in which it is studied. In one dimension, every spin has only two neighbors, and the system behaves unremarkably. However, it exhibits the aforementioned phase transitions in two or more dimensions (i.e., four or more neighbors).

While the Ising model has originally been used to model magnetism, by now there are many more applications: From protein folding \cite{isingapplbio} to social sciences \cite{isingapplsoc} and modeling the stock market \cite{isingapplstock}, by now there are many more problems that can be modeled by an Ising model.

\subsection{Science-Inspired Machine learning}
There is a deep connection between machine learning and natural sciences, such as physics. Not only are machine learning methods often used to conduct scientific research \cite{mlscibio,mlsciphys,mlscichem}, but knowledge from science is also frequently utilized to enhance machine learning methods, invent new ones, or elucidate existing ones.

Statistical physics can be used to explain the behavior of neural networks \cite{statphysNN} and has directly inspired the development of Boltzmann machines \cite{boltzmannMachine}. Particle swarm optimization \cite{particleswarm} and Hopfield networks \cite{hopfield} are inspired by the behavior of animals and neurons, respectively.

Notably, Simulated Annealing \cite{simulatedAnnealing}, a commonly used evolutionary optimization strategy inspired by metallurgical research, utilizes a temperature variable to control the randomness of a system and guide it toward an optimal state. We will use this algorithm as a competitor.

Additionally, Ising models have been utilized to comprehend the training of neural networks \cite{isingNN} and have inspired Markov random fields \cite{markovRandomFields}, which are employed to model joint probability distributions.
Still, to the best of our knowledge, they have not been used as a basis for an evolutionary algorithm.

\subsection{Ensemble methods}
Replacing a complex model with an ensemble of models can often be an effective way to enhance the model's performance and increase its reliability.
While there are many approaches to ensembles (like stacking or boosting \cite{stackingad,boosting}), the most commonly studied task combines multiple independent models using a combination function \cite{bagging}. For this to help, we require the errors of ensemble submodels to differ from each other \cite{ensembleerrorcancel}.

This difference can be achieved in one of two ways. Either by changing the setup of each submodel, or by, for example, training different types of models. However, this makes the resulting combination task harder and can often (especially for unsupervised tasks) limit the resulting performance \cite{ijcnnensemble, feature-bagging}.

So instead, submodels are often identical, except for a random seed \cite{randomforest}. However, combining this with an optimization algorithm like gradient descent \cite{gradientdescent} or evolutionary optimization \cite{evolution} means that we can only achieve the variance needed for the ensemble to work when there are many different, equally easily reachable local minima. Since this assumption is often not fulfilled and requires retraining many submodels multiple times, we instead suggest a new optimization method that can directly find a set of local minima and thus an entire ensemble during training.

\section{Ising based Evolution}\label{sec:approach}

We describe our approach in Alg. \ref{alg:ising}.
We present here the two-dimensional version, but this can be easily adapted by modifying what it means for two indices to be neighboring. While Ising-based evolution is similar to cellular evolution \cite{cellularevolution} as updates happen locally, the likelihood of changing a value is given by Eq. \eqref{eqn:prob}:
\begin{equation}
    p(old,new)=\begin{cases}
1 &\text{if } old>new\\
\exp(\beta\cdot (old-new)) &\text{else}
\end{cases}
    \label{eqn:prob}
\end{equation}
When minimizing the existing value, the probability is $p=1$, and such the update step happens. However, if it is not improved, there is still a chance that the value will get updated. This is necessary for the method to avoid getting stuck in local minima. Here, the probability is bigger when $\beta$ is smaller. This is the same probability used in the usual Ising model to go to a state of different energy.

We chose here a mutation of our value that is either sexual or asexual: Either two values are averaged, or one value is changed according to a normal distribution. Still, this mutation is not part of our algorithm and can be changed at will. We will also use an equivalent mutation step in our competitor algorithms.

\begin{algorithm}[htbp]
\caption{Ising based evolution}\label{alg:ising}
\begin{algorithmic}
\Require $f(x)$
\Require $Region,\;wid,\;hei,\;\beta,\;n_{steps}>wid\cdot hei$
\Require $uniform(Region),\;normal(mean,std)$
\For{$i<wid$}
    \For{$j<hei$}
        \State $population_{i,j} \gets  uniform(Region)$
        \State $qual_{i,j} \gets  f(population_{i,j})$
        \State $n_{steps} \gets n_{steps}-1$
    \EndFor
\EndFor
\State $solution \gets  population_{argmin(qual)}$
\State $qual_{min} \gets min(qual)$

\While{$n_{steps} >0 $}
    \State $i \gets uniform(0,wid)$
    \State $j \gets uniform(0,hei)$
    \State $\Bar{i}, \Bar{j} \gets neighbour(i,j)$

    \State $test \gets \frac{population_{i,j}+population{\Bar{i},\Bar{j}}}{2}$
    \If{$uniform(0,1)<0.5$}
        \State $test \gets normal(mean=population_{i,j},std=100)$
    \EndIf

    \State $quality_{test} \gets f(test)$

    \If{$uniform(0,1)<\exp(\beta\cdot (qual_{i,j}-qual_{test}))$}
        \State $population_{i,j} \gets test$
        \State $qual_{i,j} \gets qual_{test}$
        \If{$qual_{test}<qual_{min}$}
            \State $solution \gets test$
            \State $qual_{min} \gets qual_{test}$
        \EndIf
    \EndIf
    \State $n_{steps} \gets n_{steps}-1$

\EndWhile

\Return $solution$
\end{algorithmic}
\end{algorithm}

\subsection{Comparison algorithms}\label{sec:comparison}
We chose five different evolutionary optimization strategies to compare our algorithm to.

\textbf{Cellular evolution} \cite{cellularevolution} is a similar approach to Alg. \ref{alg:ising}. The main difference to our approach is that the probability of choosing a suboptimal solution is fixed at $10\%$. While this change may seem small, it would already be sufficient to prevent the emergence of a phase transition in the physical model.

\textbf{Simulated Annealing} \cite{simulatedAnnealing} is also a physics-inspired algorithm in which one solution is updated based on mutations, with a probability given by Eq. \eqref{eqn:prob}. As it is inspired by the cooling of a metal, the value of $\beta$ increases with the number of update steps $i$. A common way to parameterize this is $\beta=\beta_0 \cdot i$. However, to allow for more variance with a high number of update steps and have a stronger competitor, we choose $\beta=\beta_0 \cdot \sqrt{i}$.

\textbf{Mutation} evolution is a much simpler algorithm, which generates a new value through a standard distribution $ newvalue = normal(mean=current\,Value,std=100)$. The value is updated if the new value has a lower score than the old one. We employ this algorithm here, similar to an ablation study, as it utilizes a similar update step to our algorithm. Still, Mutation evolution is a case of differential evolution \cite{diffevolution}.

\textbf{Mixture} evolution is a further case of differential evolution \cite{diffevolution}, which we use like an ablation study. Here we are given a population of $100$ random values, and we randomly average the value of two instances to generate a new one. If this generated value outperforms one of its parents, or in at least 10\% of cases, we keep it. While Mutation evolution could be seen as inspired by asexual reproduction, Mixture evolution is more similar to sexual reproduction.

\textbf{Random Search} is our final and simplest algorithm. Instead of using a procedure to update the current value, we randomly guess it. While this algorithm cannot get stuck in local minima, it can also not benefit from any local feedback. The likelihood of reaching the optimal state is the same, whether in the second-best state or in the worst one.

\section{Test Description}\label{sec:method}

To test how well a particular algorithm performs in finding the global minimum of a function, we need to evaluate it on a function with many local optima. To achieve this, we suggest minimizing Equation \eqref{eqn:thefunc}.

\begin{equation}
    f(x)=\left|\sin\left(\frac{x+1}{100}\right)\right|,\quad x\in \mathbb{Z}
    \label{eqn:thefunc}
\end{equation}

\begin{figure}[htbp]
\centerline{\includegraphics[width=0.95\columnwidth]{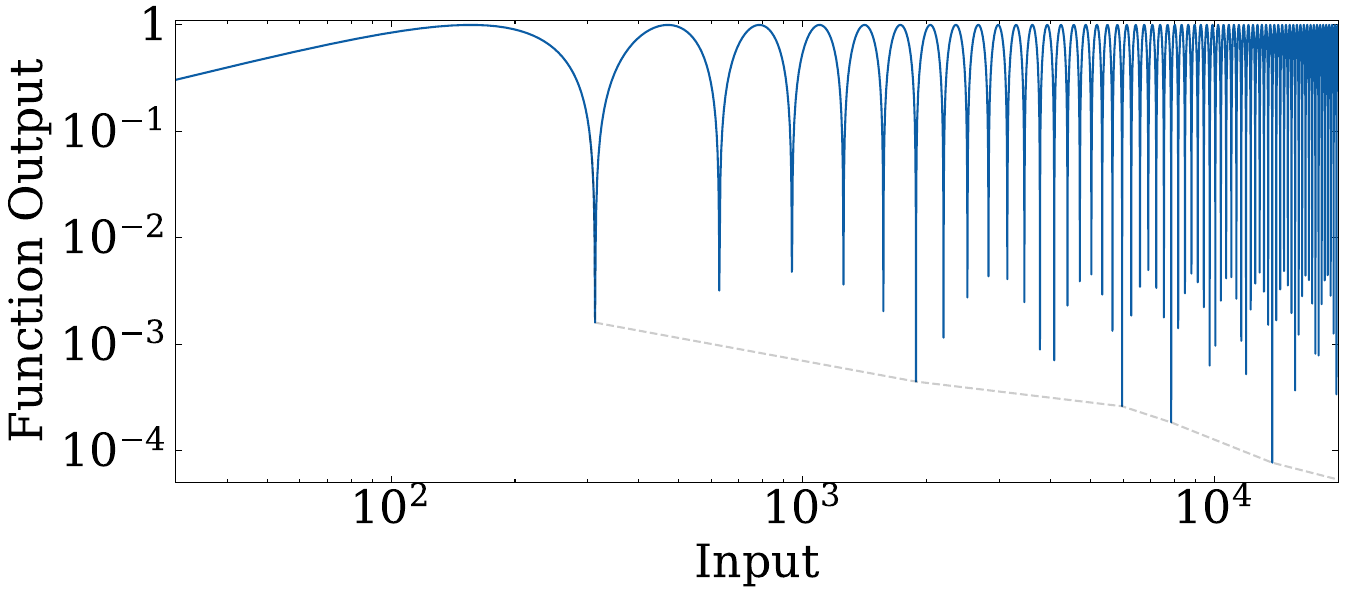}}
\caption{Visualization of Eq. \eqref{eqn:thefunc}. Because of the double logarithmic axes, the sinus looks unusual. But this highlights that the local minima are not randomly distributed.}
\label{fig:func}
\end{figure}

We visualize this function in Fig. \ref{fig:func}.

While Eq. \eqref{eqn:thefunc} is almost a simple sine function, the evaluation over the range of integers means that most minima are only local minima (See Theorem \ref{theo:glob}). Still, the periodicity and improving approximation of a real number impart the function with certain general properties. For example $f(x)\approx 0 \rightarrow f(n\cdot x)\approx 0\; \forall n \in \mathbb{N}$.

\begin{theorem}
The only global optimum to function \eqref{eqn:thefunc} is $x=-1$.
\label{theo:glob}
\end{theorem}
\begin{proof}
\begin{equation*}
    f(-1)=\left|\sin(0)\right|=0;\quad f(x)\geq 0\; \forall x
\end{equation*}
\begin{equation*}
    f(x)=0, x\neq -1 \implies \frac{x+1}{100} = k\cdot \pi, k \in \mathbb{Z} \setminus \{0\}
\end{equation*}
\begin{equation*}
    \implies \pi = \frac{x+1}{100\cdot k} \in \mathbb{Q} \quad \text{which contradicts that } \pi \text{ is irrational.}
    \end{equation*}
\end{proof}

Thus, we can exclude the global minimum by only considering $x \in \mathbb{N}$. We also only evaluate $x<10^5$ to ensure an optimal result and to put evaluation costs into perspective. In total, this means there are $318$ local minima of our function.

\begin{figure}[htbp]
\centerline{\includegraphics[width=0.95\columnwidth]{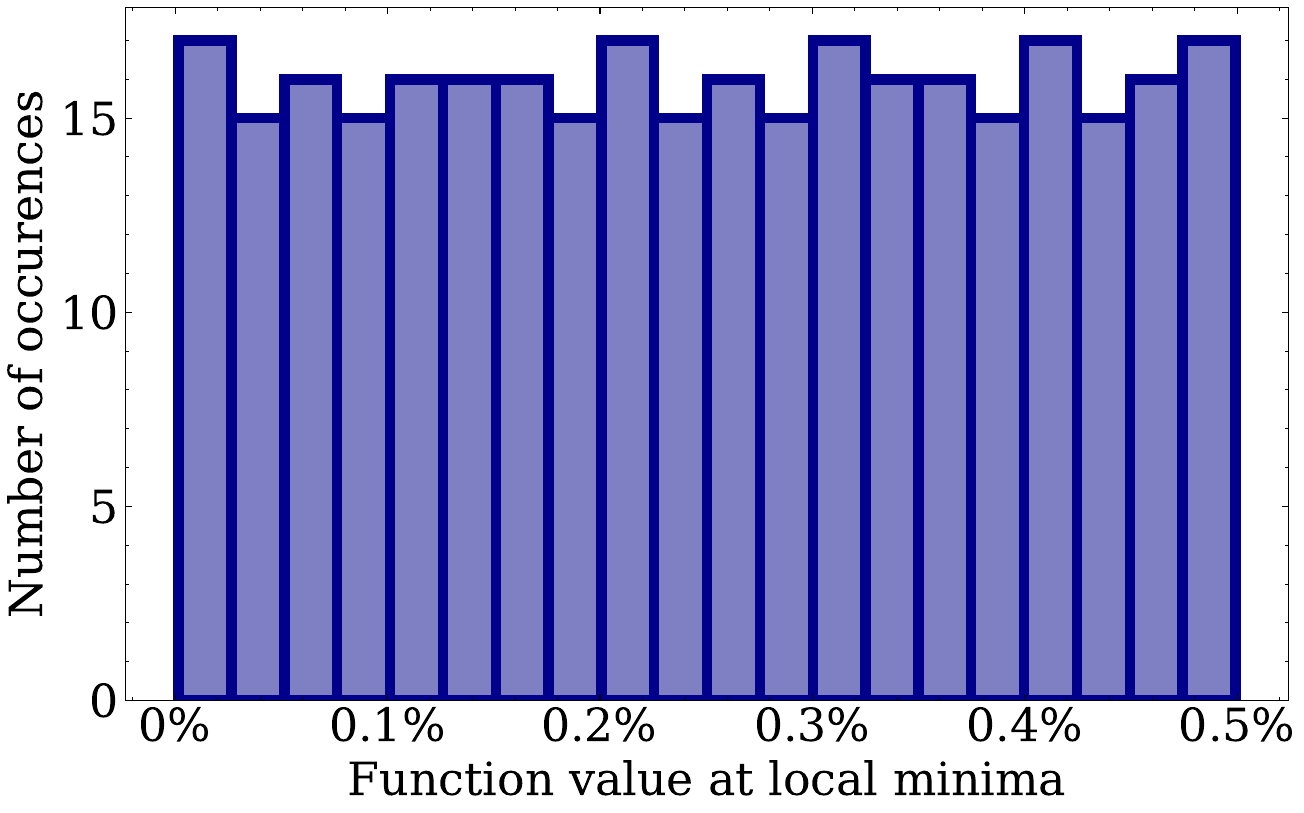}}
\caption{Distribution of locally minimal function values.}
\label{fig:dist}
\end{figure}

Further, as shown in Fig. \ref{fig:dist}, the quality of minima is equally distributed. This means, we can use the average found function value, to evaluate an optimization algorithm: When an algorithm only finds a local minimum, its average performance will be about $0.25\%=2.5\cdot 10^{-3}$. A more efficient algorithm would instead find the lowest point of our range, $f(84822) \approx 1.64 \cdot 10^{-5}$.

\section{Results}\label{sec:results}

Using this setup, we randomly initialize and run each algorithm $100$ times, measuring the average result found. Each algorithm can evaluate the function up to $10^5$ times to find the lowest function value. In theory, every function value in our range could be evaluated, guaranteeing that the lowest value is found. Still, no algorithm reliably finds the lowest performance all the time since we allow our algorithm to make the same function call multiple times. While repetitions should usually be cached in an actual application, this implicitly allows us to punish algorithms that do not efficiently explore the entire search space and guarantees that algorithms run in finite time.

\begin{figure}[htbp]
\centerline{\includegraphics[width=0.95\columnwidth]{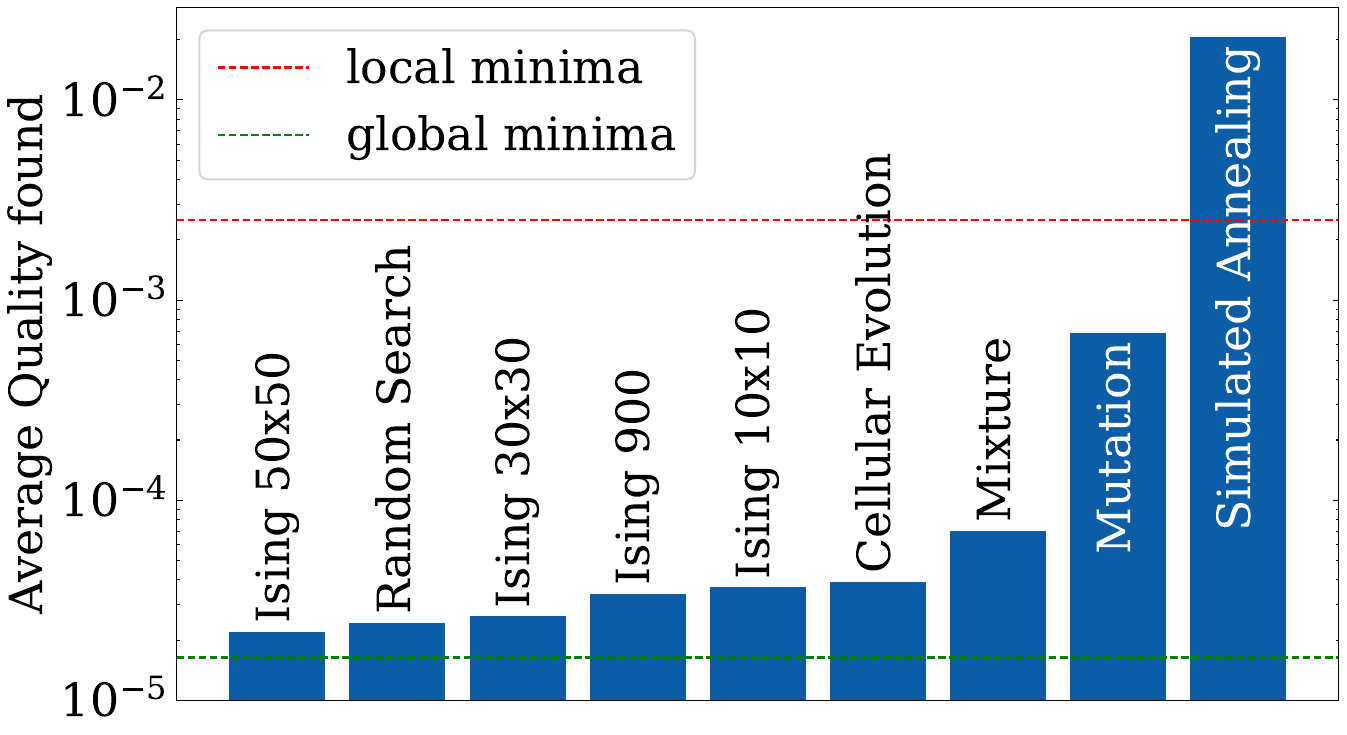}}
\caption{Average minimal value found for different evolutionary optimization algorithms.}
\label{fig:really}
\end{figure}

The resulting average function values are shown in Fig. \ref{fig:really}.

The Mutation algorithm does not improve much upon finding a local minimum. Simulated Annealing also freezes quickly. Interestingly, the Mixture algorithm does much better than both. This is likely because if $f(x)\approx 0 \approx f(y)$ this implies that $f(\frac{x+y}{2})\approx 0$ or $f(\frac{x+y}{2})\approx 1$, which is a property we can not expect to generalize to different tasks.

Each of our Ising-based models performs much better, with the best reaching an average performance of $\approx 2.2\cdot 10^{-5}$, close to the theoretical optimum.
We will spend the remainder of this section to further understand this concept.
Importantly, our Ising-based optimization outperforms the related Cellular Evolution.
We used $\beta=100$ here, but choosing a different value does not significantly affect the resulting quality, up to the limit studied in Section \ref{sec:phase}.

The Random Search model seems to be the most robust competitor algorithm. This is a result of our construction: The likelihood that it does not find the optimal value is $(1-\frac{1}{10^5})^{10^5}\approx 0.3679\approx \frac{1}{e}$, and we could improve it further by removing duplicate function calls.
Still, while it can find absolute minima relatively commonly, it cannot benefit from local feedback. This means its practical applications are limited, as most practical functions do not contain as many local minima, and a higher number of optimizable parameters exponentially increases the number of samples to be tried.

\begin{figure}[htbp]
\centerline{\includegraphics[width=0.95\columnwidth]{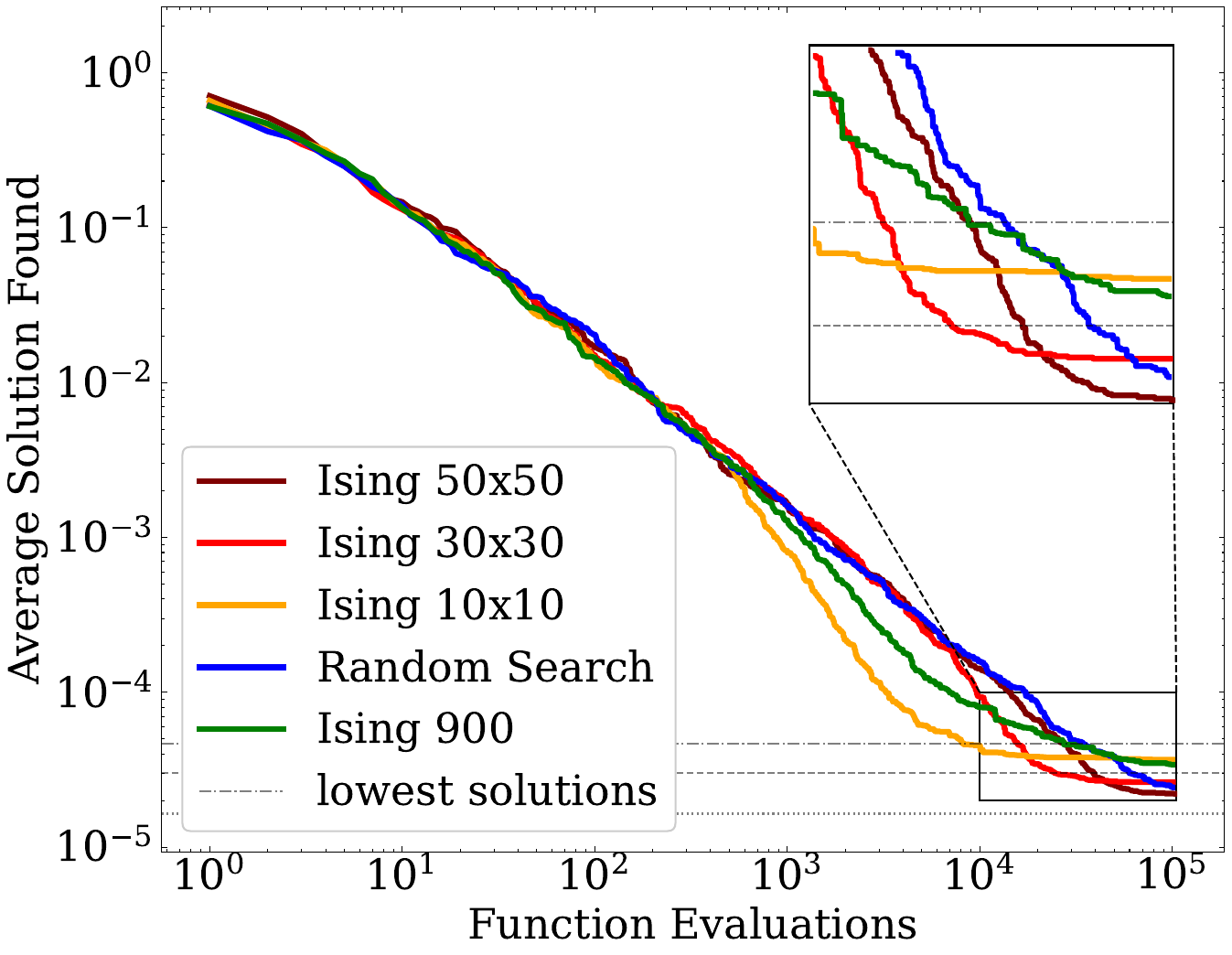}}
\caption{Average lowest value found as a function of the number of function evaluations used. We zoom in on this region because the most interesting changes happen in the lower right corner. We also add the lowest three possible solutions of Eq. \eqref{eqn:thefunc} to be found as horizontal lines.}
\label{fig:hereswhy}
\end{figure}

We study this further by comparing the performance of our Ising-based approach with that of Random Search as a function of the number of function evaluations allowed. This is shown in Fig. \ref{fig:hereswhy}.

It seems that for the first function evaluations, each algorithm shown performs very similarly. This is because the population is first initialized randomly, making the Ising-based algorithm equivalent to the Random Search for $10^2$ to $25^2\approx 10^{2.8}$ function evaluations.
Afterward, each Ising-based algorithm decreases more quickly than the Random Search algorithm, which means that they are faster in reaching a better solution than the Random Search and demonstrate the benefit of local feedback in our algorithms. The change might look small, but due to the double logarithmic axes, even a small change can result in a tenfold speed increase.
But the model performance converges at some point and does not improve further. Both the resulting quality and the cost of evaluations seem to be a function of the size of the population of our algorithm: the bigger the board, the longer we follow the Random Search performance, but the lower the resulting quality is also. And ultimately, there is a point where a 50x50 population converges at a quality below Random Search in $\approx 50\%$ fewer function evaluations.

As an ablation study, we also show a one-dimensional Ising-based model with the same population size of the 30x30 two-dimensional model. However, it appears to converge more slowly and exhibit poorer performance than its two-dimensional alternative. This is expected from the behaviour of the physical Ising model and will be further understood in the following subsections.

\subsection{Phase transitions}\label{sec:phase}

The Ising model is often studied in physics because it is one of the simplest models that can show phase transitions. While for high temperatures (low $\beta$), the system is chaotic and almost random, for low temperatures (high $\beta$), relatively stable regions of aligned spins are formed. Interestingly, this effect does not appear for the one-dimensional Ising model but only in at least two dimensions. We aim to investigate whether a similar transition also occurs here.

\begin{figure}[htbp]
\centerline{\includegraphics[width=0.95\columnwidth]{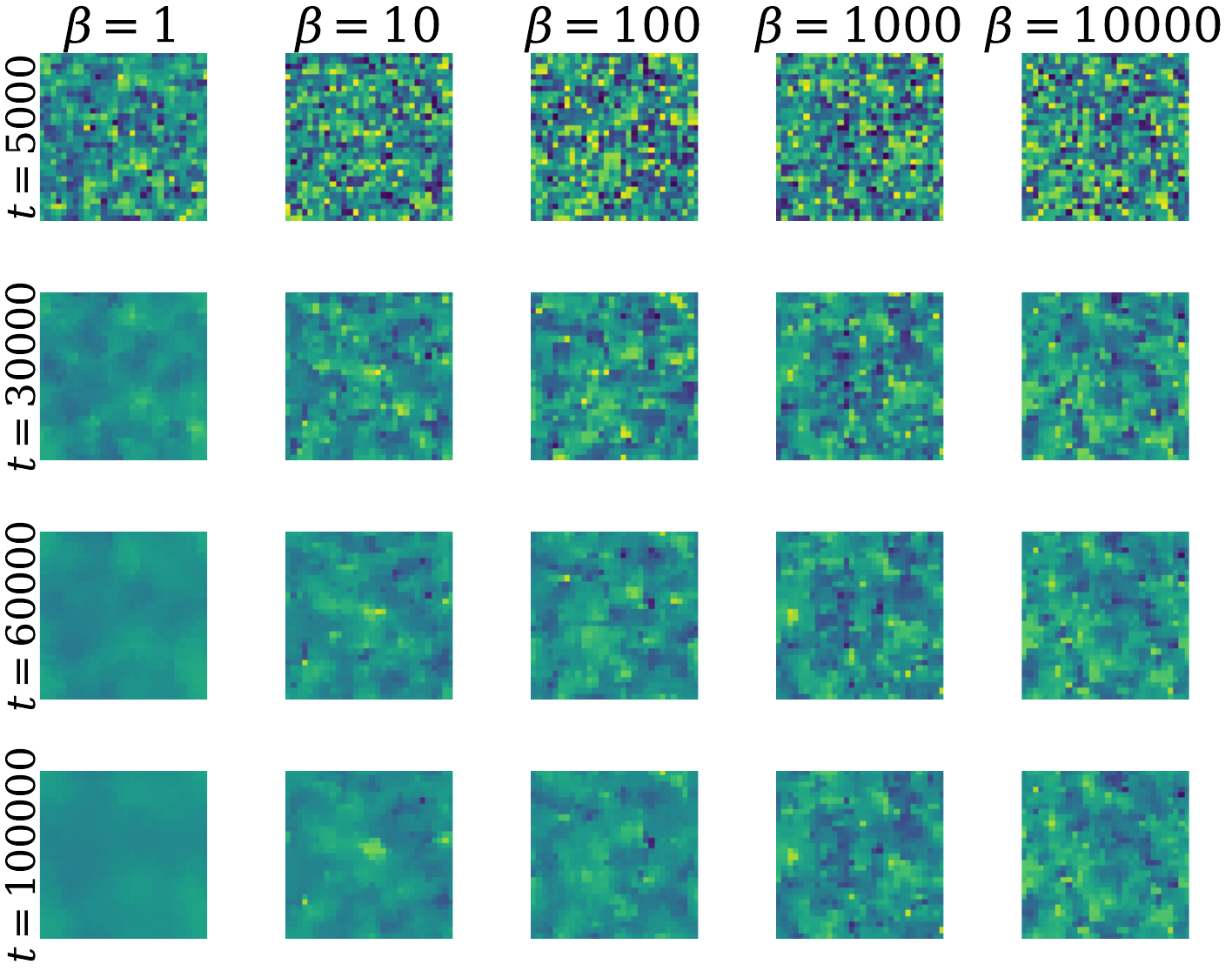}}
\caption{Current state (color) over the population, as a temperature and update step function.}
\label{fig:phase}
\end{figure}

To illustrate this, we present the current state of our population for various temperatures after different numbers of update steps for the 30x30-sized algorithm in Fig. \ref{fig:phase}.

Similar to the physical effect, we see a relatively stable distribution. After the almost random initialization, each model exhibits unique structures, which persist even after further update steps, albeit with slight changes (for example, the yellow spot in the center of $\beta=10$).

Similarly, this distribution also depends on the temperature value, just like the physical effect.
For a high temperature (low $\beta$), the resulting structures seem to disappear the most with increasing update steps. However, for low temperatures (high $\beta$), the results become more chaotic, and stable regions emerge.

But after $10^5$ update steps, even the almost convergent distribution of $\beta=1$ still differs by about $\approx 15\%$ of the available range in its solutions.

\begin{figure}[htbp]
\centerline{\includegraphics[width=0.95\columnwidth]{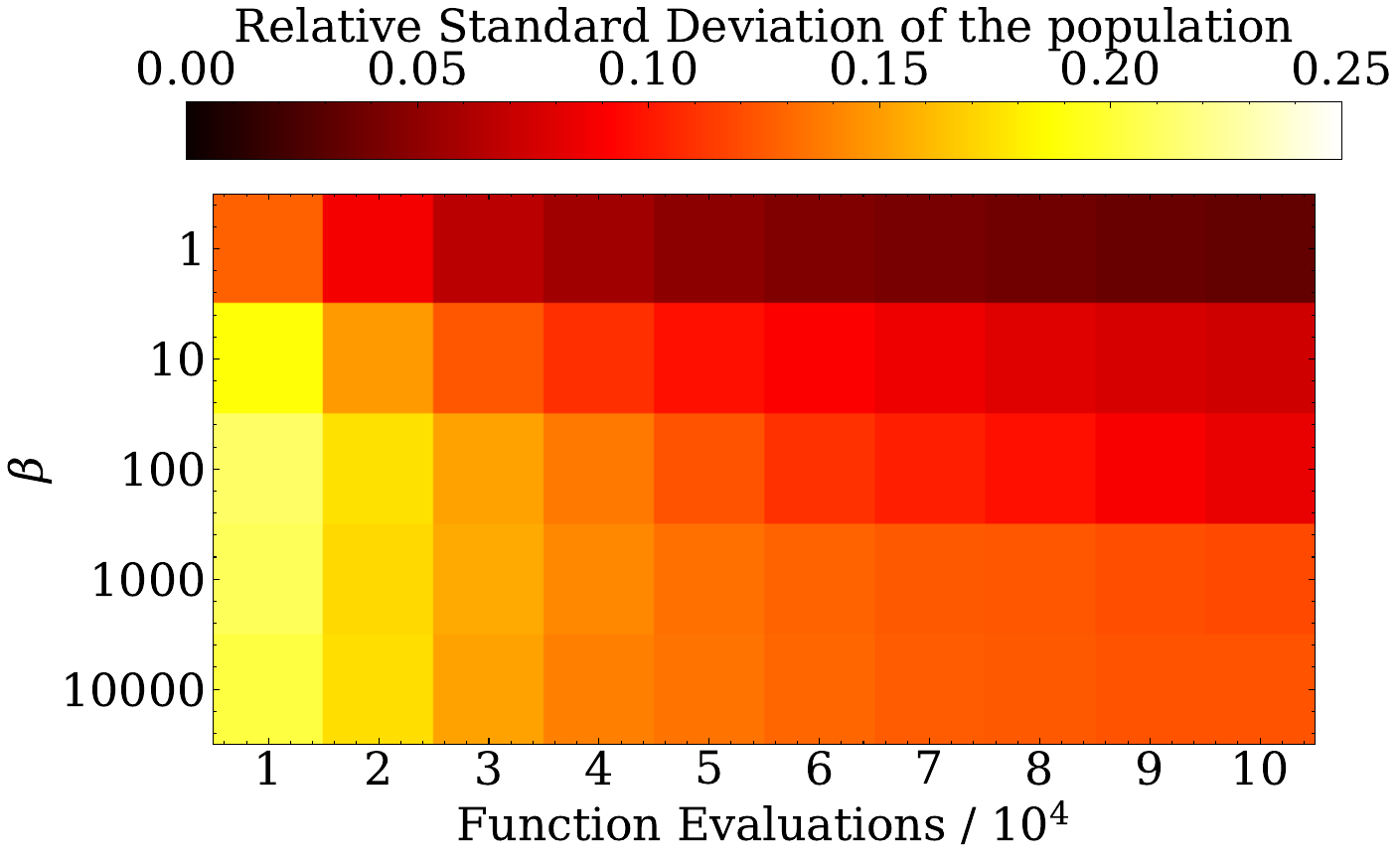}}
\caption{Relative standard deviation (standard deviation divided by $10^5$) of models with different $\beta$ value during the optimization process for a two-dimensional ising model of population size 30x30.}
\label{fig:apotheosis}
\end{figure}

\begin{figure}[htbp]
\centerline{\includegraphics[width=0.95\columnwidth]{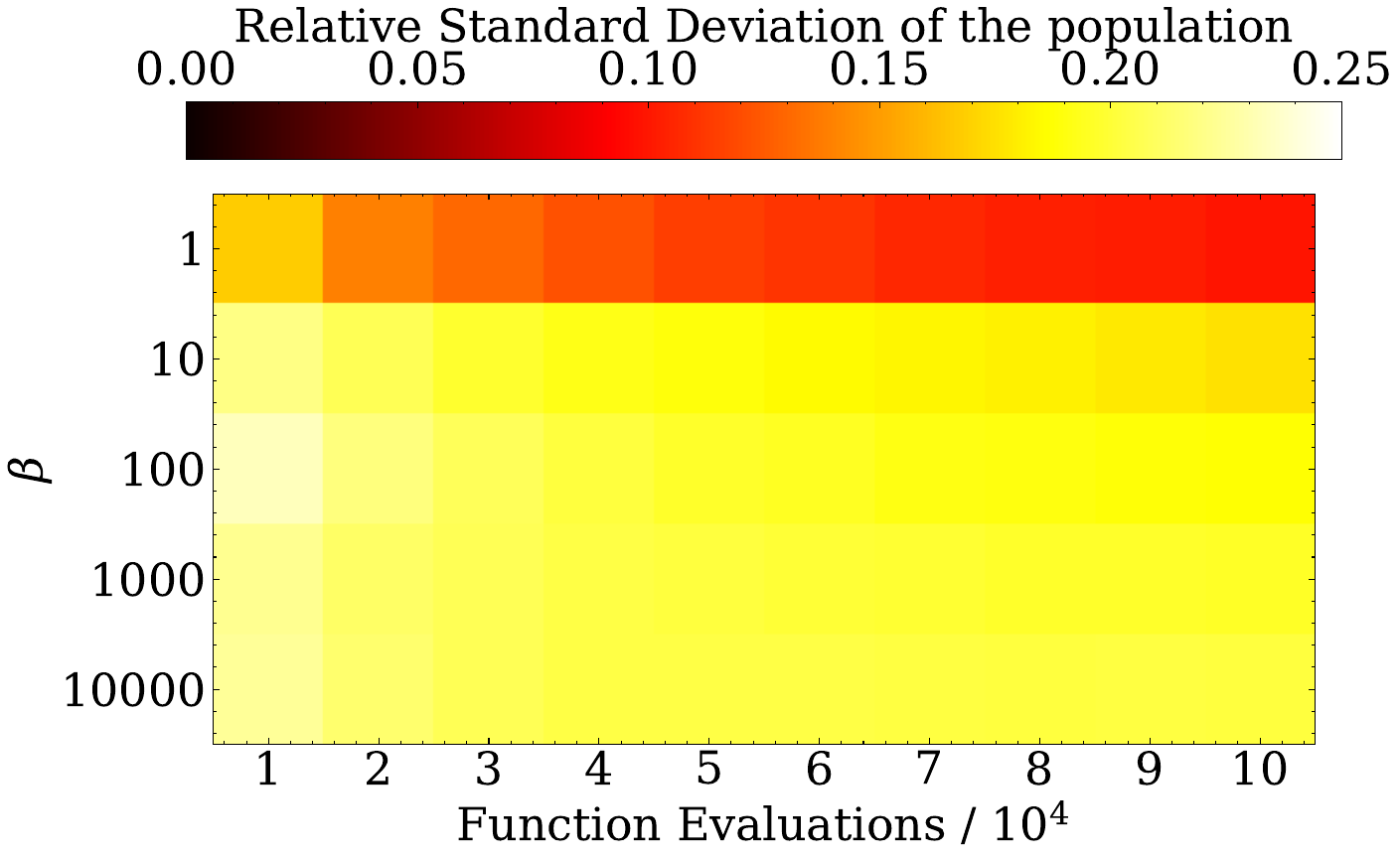}}
\caption{Relative standard deviation (standard deviation divided by $10^5$) of models with different $\beta$ value during the optimization process for a one-dimensional ising model of population size 900.}
\label{fig:linearity}
\end{figure}

To further characterize how the distribution depends on temperature and the update step, we visualize the standard deviation of these images as a function of both. We show this for a two-dimensional model in Fig. \ref{fig:apotheosis} and a one-dimensional one in Fig. \ref{fig:linearity}.

As we have seen before, in the two-dimensional case, the system becomes more chaotic with a higher value of $\beta$, while also needing a longer time to converge. We also almost see phases: For a low number of evaluations and a high value of $\beta$, the variance is high, while for a high number of evaluations and a low $\beta$ it is significantly smaller. However, importantly, the population variance does not drop to zero, unlike the other evolution algorithms. This is not only an effect of continuing random mutation steps, as these should at most create a relative standard deviation of $\frac{100}{10^5}=0.001$ (See Alg. \ref{alg:ising}).

Interestingly, and similarly to the physical effect, the one-dimensional case appears to behave quite differently from the two-dimensional one. While the two-dimensional model shows a gradual transition from the low to the high variance phase, the one-dimensional one freezes out almost completely for every $\beta>1$.

\subsection{Ensembles}

\begin{figure}[htbp]
\centerline{\includegraphics[width=0.95\columnwidth]{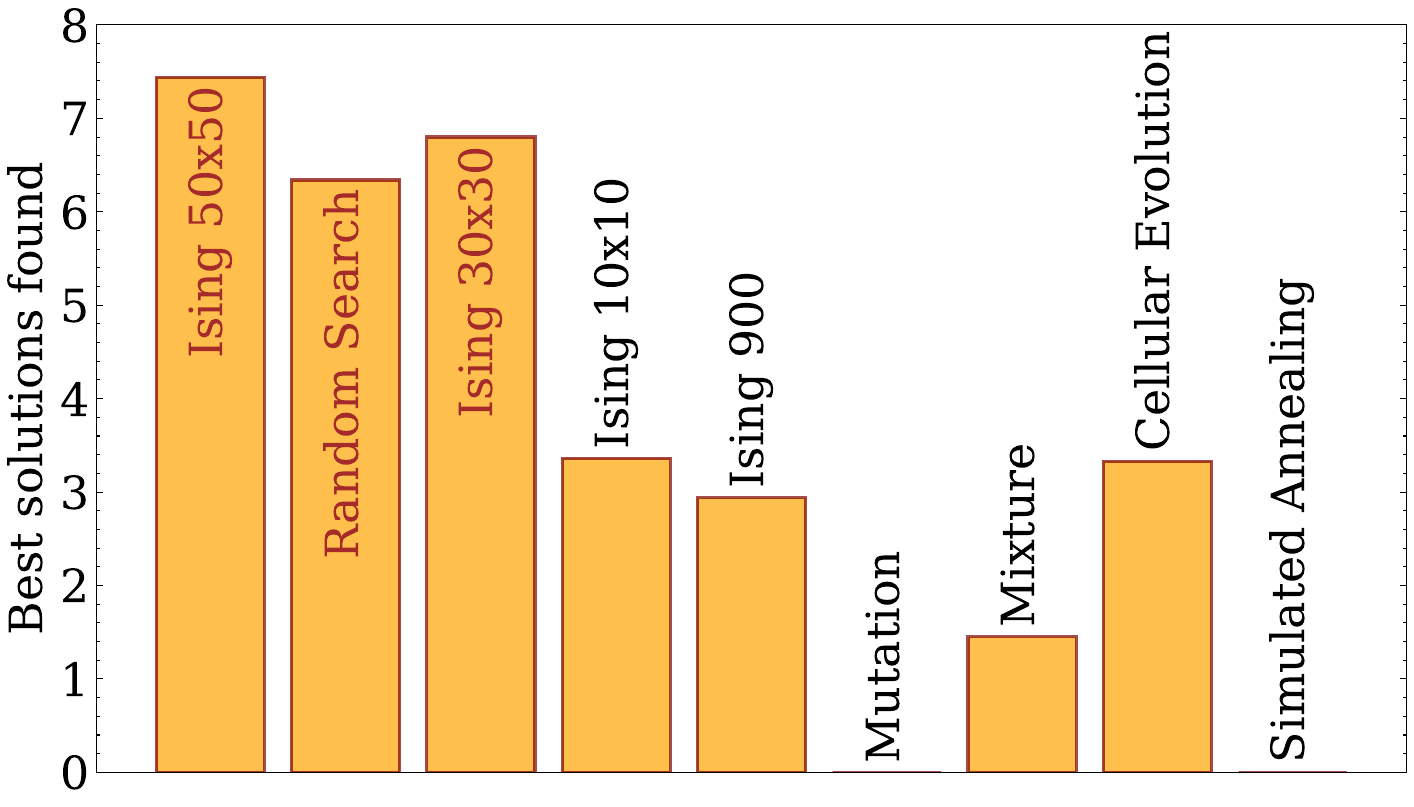}}
\caption{When not only searching for the best solution but also for the 9 best local optima, how many of these are found on average by a given method.}
\label{fig:nsemble}
\end{figure}

Finally, we aim to investigate how effectively our Ising-based approach can be utilized to identify not only the global minimum of a function but also multiple distinct ones that can be combined into an ensemble. To do this, we test not how often the global minimum is found but how many of the lowest ten solutions are found. This is shown in Fig. \ref{fig:nsemble}.

Our Ising-based approach identifies most of these minima, with an average of $ 7.44$ out of $ 10$ valuable minima being found. Similar to our analysis in Fig. \ref{fig:really}, a larger size of the Ising population appears to be beneficial, and here the improvement over the Random Search algorithm increases, as even a 30x30 population is sufficient to outperform it.

This demonstrates that utilizing submodels from various regions can yield ensembles that outperform those randomly initialized. We can create an ensemble from our results by requiring only one optimization loop, rather than multiple ones (one per submodel). Additionally, we could link the number of different solution regions to the number of ensemble submodels needed. But this still requires further study.

\section{Conclusion}\label{sec:conclusion}
This paper introduced an evolutionary optimization algorithm based on Ising models.
Our method is capable of finding global minima even in highly complex loss functions.
It achieves this by leveraging knowledge from theoretical physics, adapting a model that demonstrates the emergence of semi-stable regions, and then adjusting these regions to evolutionary optimization.
We have demonstrated that our algorithm exhibits properties similar to those that make Ising models so interesting in physics.

We can utilize these regions to perform what we call high-variance optimization. This kind of optimization allows us not only to find better optima but also has interesting applications for ensembles.

In further studies, we aim to characterize other shapes of our population (higher dimensionality, non-square shapes) and investigate whether we can improve the trade-off between quality and time cost by using a region-based initialization.
Applying this method to various other loss functions and mutation procedures would also be interesting.

Further, we are interested in the possibility of learning complicated and inherently explainable models (like a function tree) using an algorithm like ours.

\bibliographystyle{IEEEtran}
\bibliography{refs,new}

\end{document}